\documentclass[runningheads]{llncs}
\usepackage{graphicx} % Required for inserting images

\title{Multimodal Deep Learning for Prediction of Progression-Free Survival in Patients with Neuroendocrine Tumors Undergoing \textsuperscript{177}Lu-based Peptide Receptor Radionuclide Therapy}
\titlerunning{Multimodal Deep Learning for PFS prediction in NET Patients} 

\usepackage{graphicx}
\usepackage{booktabs}
\usepackage{amssymb}
\usepackage{caption}
\usepackage{subcaption}
\usepackage{tabularx}
\usepackage{hyperref}
\usepackage{float}
\usepackage{amsmath}
\usepackage{placeins}   
\usepackage{booktabs}
\usepackage{tabularx}
\usepackage[utf8]{inputenc}
\usepackage{array} 
\usepackage{comment}
\usepackage{subcaption}

\author{Simon Baur\inst{1} \and
Tristan Ruhwedel\inst{2} \and
Ekin Böke\inst{1} \and
Zuzanna Kobus\inst{3,4} \and
Gergana Lishkova\inst{5} \and
Christoph Wetz\inst{2} \and
Holger Amthauer\inst{2} \and
Christoph Roderburg\inst{6} \and
Frank Tacke\inst{3} \and
Julian M. Rogasch\inst{2} \and
Wojciech Samek\inst{1,7,8} \and
Henning Jann\inst{3} \and
Jackie Ma\inst{1} \and
Johannes Eschrich\inst{3,9}}

\institute{Department of Artificial Intelligence, Fraunhofer Heinrich Hertz Institute, 10587 Berlin, Germany 
\and
Department of Nuclear Medicine, Charité—Universitätsmedizin Berlin 
\and
Department of Hepatology and Gastroenterology, Charité—Universitätsmedizin Berlin 
\and
Division of Interventional Radiology, Department of Radiology, Memorial Sloan Kettering Cancer Center
\and
Department of Endocrinology and Metabolism, Charité—Universitätsmedizin Berlin
\and
Clinic for Gastroenterology, Hepatology and Infectious Diseases, University Hospital Düsseldorf, Medical Faculty of Heinrich Heine University Düsseldorf 
\and
BIFOLD $-$ Berlin Institute for the Foundations of Learning and Data 
\and
Department of Electrical Engineering and Computer Science, Technische Universität Berlin 
\and
Berlin Institute of Health at Charité – Universitätsmedizin Berlin }

\authorrunning{Baur, Eschrich et al.}
\begin{document}
\maketitle

\begin{abstract}
Peptide receptor radionuclide therapy (PRRT) is an established treatment for metastatic neuroendocrine tumors (NETs), yet long-term disease control occurs only in a subset of patients. Predicting progression-free survival (PFS) could support individualized treatment planning. This study evaluates laboratory, imaging, and multimodal deep learning models for PFS prediction in PRRT-treated patients.

\textbf{Methods} In this retrospective, single-center study 116 patients with metastatic NETs undergoing [\textsuperscript{177}Lu]Lu-DOTATOC were included. Clinical characteristics, laboratory values, and pretherapeutic somatostatin receptor positron emission tomography/computed tomographies (SR-PET/CT) were collected. Seven models were trained to classify low- vs. high-PFS groups, including unimodal (laboratory, SR-PET, or CT) and multimodal fusion approaches. Performance was assessed via repeated 3-fold cross-validation with area under the receiver operating characteristic curve (AUROC) and area under the precision-recall curve (AUPRC). Explainability was evaluated by feature importance analysis and gradient maps.

\textbf{Results} Forty-two patients (36\%) had short PFS ($\leq$ 1 year), 74 patients long PFS ($>$1 year). Groups were similar in most characteristics, except for higher baseline chromogranin~A ($p = 0.003$), elevated $\gamma$-GT ($p = 0.002$), and fewer PRRT cycles ($p < 0.001$) in short-PFS patients. The Random Forest model trained only on laboratory biomarkers reached an AUROC of $0.59 \pm 0.02$. Unimodal three-dimensional convolutional neural networks using SR-PET or CT performed worse (AUROC $0.42 \pm 0.03$ and $0.54 \pm 0.01$, respectively). A multimodal fusion model laboratory values, SR-PET, and CT -augmented with a pretrained CT branch - achieved the best results (AUROC $0.72 \pm 0.01$, AUPRC $0.80 \pm 0.01$).

\textbf{Conclusion} Multimodal deep learning combining SR-PET, CT, and laboratory biomarkers outperformed unimodal approaches for PFS prediction after PRRT. Upon external validation, such models may support risk-adapted follow-up strategies.
\end{abstract}

\section{Introduction}
Neuroendocrine tumors (NETs) arise from  neuroendocrine cells and represent a heterogeneous group of neoplasms with variable biological behavior and clinical presentation \cite{pedraza2018multilayered}. Although classified as rare, the reported incidence of NETs has been steadily increasing over recent decades \cite{Liu2024}. Most frequently, NETs originate in the gastrointestinal tract or pancreas, collectively referred to as gastroenteropancreatic neuroendocrine tumors (GEP-NETs) \cite{Yao2008OneHY}.  In a subset of patients, the primary tumor site remains unknown despite extensive diagnostic work-up, referred to as NETs of unknown primary (CUP-NETs) \cite{hainsworth2024neuroendocrine}. For patients with advanced disease, available treatment options are limited and include somatostatin analogues, targeted therapies, chemotherapy, and peptide receptor radionuclide therapy (PRRT). PRRT with [\textsuperscript{177}Lu]Lu-DOTATATE or [\textsuperscript{177}Lu]Lu-DOTATOC has emerged as an effective treatment strategy for patients with metastatic NETs that express high levels of somatostatin receptors \cite{pellat2021neuroendocrine}. Clinical trials have demonstrated that \textsuperscript{177}Lu-based PRRT significantly prolongs progression-free survival (PFS) \cite{strosberg2017phase}. More recently, the NETTER-2 trial evaluated [\textsuperscript{177}Lu]Lu-DOTATATE as a first-line therapy for patients with advanced grade 2 and 3 gastroenteropancreatic (GEP) NETs, demonstrating encouraging outcomes that support its expanded role in earlier lines of treatment \cite{singh2024177lu}. Nonetheless, meta-analyses have demonstrated that PRRT achieves objective response rates in patients with advanced NETs ranging between 25.0\% and 35.0\%, depending on the response assessment criteria applied, and disease control rates between 79.0\% and 83.0\% \cite{pmid32118772}. Identifying patients who will not achieve long-term disease control or remission in advance is a clinical need and defines the rationale of the present study. As of today, histological Ki-67 proliferation index and serum chromogranin A (CgA) remain the most established prognostic biomarkers in patients undergoing PRRT. Elevated Ki-67 and CgA levels have been consistently associated with shorter PFS \cite{daskalakis2024modified}. The multigene transcriptomic assay NETest has been proposed as a predictive tool for PRRT outcomes, showing promising initial results, however, its high cost and limited availability currently restrict clinical implementation \cite{knigge2017enets}. Recent work by Ruhwedel et al. identified the De Ritis ratio (AST/ALT) as a prognostic biomarker in patients undergoing PRRT, with elevated values associated with shorter progression-free survival and overall survival \cite{10.3390/cancers13040635,ruhwedel2025beyond}.Imaging-derived parameters have also been proposed to predict PRRT outcomes. Somatostatin receptor (SR) heterogeneity, high lesional SR expression - as assessed by the Krenning score - and the metastases-to-liver ratio (M/L ratio) have been investigated as potential predictors of therapy response \cite{werner2016survival,wetz2019association}. Furthermore, radiomic signatures extracted from baseline somatostatin receptor PET (SR-PET) or CT imaging have demonstrated potential for stratifying patients \cite{laudicella202268ga}. However, the predictive value of these parameters remains limited, thus restricting their clinical applicability. In recent years, artificial intelligence (AI) – and particularly deep learning (DL) – has emerged as a potent tool to extract high-dimensional patterns from heterogeneous biomedical data, including imaging, genomics, and clinical variables. In oncology, multimodal DL approaches leverage specialized architectures that process each data type in dedicated branches before combining them into a joint representation for prediction. Typically, convolutional neural networks are employed for image analysis, while feed-forward networks handle structured clinical data, and sequence models such as recurrent or transformer architectures are used for genomic or temporal inputs. These modality-specific encoders are integrated via fusion strategies ranging from simple concatenation to attention-based transformers enabling end-to-end optimization across all modalities \cite{acosta2022multimodal,baur2025effectiveness}. Recent reviews highlight that such architectures can capture complementary information, mitigate modality specific biases, and improve generalization across diverse patient populations \cite{acosta2022multimodal,rajpurkar2022ai_health}. This paradigm is particularly relevant for PRRT, where reliable biomarkers for predicting durable response remain scarce. Applying multimodal AI frameworks to NETs could therefore facilitate more accurate patient selection and personalized therapeutic strategies.

\section{Material and Methods}
\subsection{Patient Cohort}
This retrospective, single-center study included 116 consecutive patients with histologically confirmed NETs who received PRRT with [\textsuperscript{177}Lu]Lu-DOTATOC between 2015 and 2022 at Charité—Universitätsmedizin Berlin. Eligibility criteria comprised: (1) metastatic, progressive disease; (2) sufficient SR expression confirmed by pretherapeutic [\textsuperscript{68}Ga]Ga-DOTATOC PET/CT; (3) availability of both pretherapeutic laboratory data and imaging (SR-PET and CT scan); and (4) availability of clinical follow-up data.
Baseline laboratory values had to include liver function parameters - namely aspartate transaminase (AST), alanine transaminase (ALT), and gamma-glutamyl transferase (GGT) - as well as the neuroendocrine tumor marker chromogranin A (CgA), all measured within four weeks before initiation of PRRT. Patients were excluded if they had undergone prior PRRT, had incomplete clinical records, or insufficient follow-up to assess disease progression. 
A majority of the patient cohort analyzed in this study was previously included in earlier publications \cite{10.3390/cancers13040635,wetz2023plasma}. The present study comprises additional, more recently treated patients. Furthermore, it differs methodologically by employing a multimodal deep learning framework that integrates laboratory values and imaging data (SR-PET/CT) for predictive modeling of PFS. Thus, while the patient cohort overlaps with previous studies, the methodological approach of the current work is distinct.
Table ~\ref{tab:patient_characteristics} illustrates all patient characteristics. 

\subsection{Imaging Characteristics}
Prior to PRRT initiation, all patients underwent a pretherapeutic [\textsuperscript{68}Ga]Ga-DOTA-based PET. The median interval between the pretherapeutic SR-PET and initiation of the first PRRT cycle was 36 days (IQR 44; Q1 15.5 -- Q3 59.5 days). 
PET/CT examinations were performed in our center with either a Philips Gemini TF 16 scanner with time-of-flight capability and a 16-row CT scanner \cite{surti2007performance} or a GE Discovery MI scanner with silicon photomultipliers and time-of-flight capability and a 64-row CT scanner \cite{vandendriessche2019performance}. The CT scans included in our model were exclusively those acquired simultaneously with the pretherapeutic SR-PET to guarantee spatial and temporal alignment between anatomical and functional imaging data. 
Among the 116 patients included, 74 underwent whole-body contrast-enhanced CT, while the remaining patients received whole-body non-contrast CT. 
The contrast-enhanced CT images were acquired during the venous contrast phase with a slice thickness of 3 mm. 
We deliberately used whole-body CTs to ensure that all lesions detected on SR-PET could be anatomically correlated with the corresponding CT scan across the entire field of view.

\subsection{Peptide Receptor Radionuclide Therapy and Response Assessment}
Patients received [\textsuperscript{177}Lu]Lu-DOTATOC PRRT with a median of 3 cycles (range: 1–7), each administered at a standard dose of 200 mCi (7.40 GBq). Treatment cycles were scheduled at intervals of 10 to 12 weeks. Interim response assessment was performed using [\textsuperscript{68}Ga]Ga-DOTATOC PET/CT after every two cycles, with the first evaluation following the second treatment cycle. To minimize the risk of misinterpreting radiogenic edema as disease progression (pseudo-progression), interim staging was conducted at least two months after the most recent PRRT cycle \cite{brabander2017pitfalls}. Disease progression was determined by an interdisciplinary tumor board. In patients showing progressive disease, no additional PRRT cycles were administered. Following completion of therapy, patients underwent routine follow-up imaging every 3 to 6 months. Morphological evaluation was primarily based on CT.

\subsection{Progression-Free Survival}
PFS was defined as the time from the initiation of the first PRRT cycle until the date of documented disease progression or death from any cause. Disease progression was assessed according to RECIST 1.1 criteria as determined by the local interdisciplinary tumor board. Patients without a progression event were not included in the analysis. 
For the purpose of this study, we defined the PFS threshold at 1 year, as progression within the first year after PRRT initiation is generally considered to indicate insufficient therapeutic benefit and has been used previously in clinical studies \cite{Baudin2022OCLURANDOM}. 
This threshold was applied to dichotomize patients into low-PFS ($\leq$ 1 year) and high-PFS ($>$ 1 year) groups for subsequent analyses.

\subsection{Deep Learning Models}
\begin{figure}[!h]
    \centering
    \includegraphics[width=\textwidth]{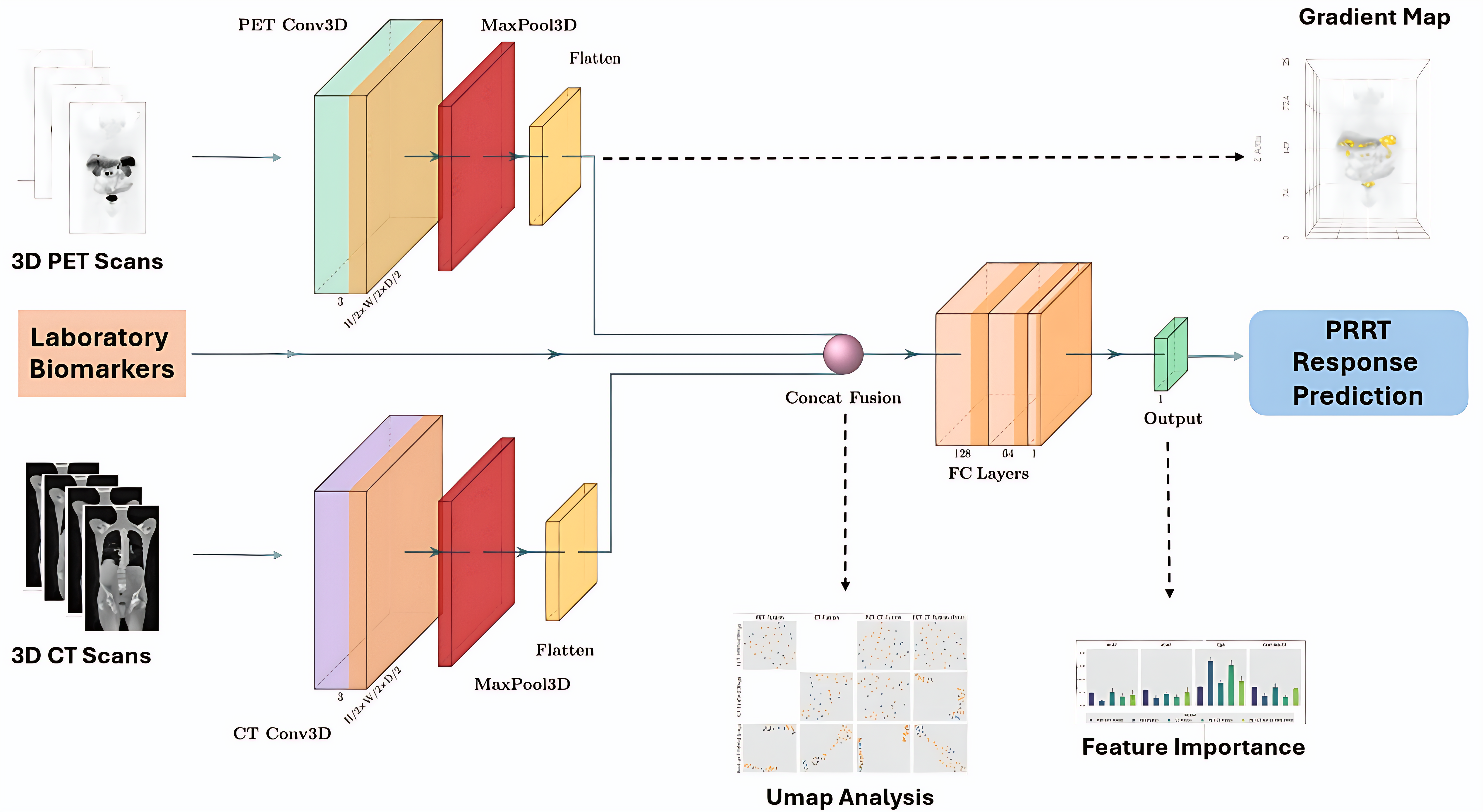}
    \vspace{0.5em} % small vertical space between image and table
    \begin{tabular*}{\textwidth}{@{\extracolsep{\fill}} c c c l}
        \toprule
        \textbf{PET} & \textbf{CT} & \textbf{Laboratory values} & \textbf{Input Setup Description} \\
        \midrule
                   &            &  \checkmark & Laboratory values only \\
        \checkmark &            &            & PET only \\
                  & \checkmark &            & CT only \\
        \checkmark &            & \checkmark & PET Fusion \\
                  & \checkmark & \checkmark & CT Fusion \\
        \checkmark & \checkmark & \checkmark & PET CT Fusion \\
        \checkmark & \checkmark & \checkmark & PET CT Fusion (pretrained CT)\\

        \bottomrule
    \end{tabular*}
    \caption{(Top) Overview of the proposed deep learning pipeline for PRRT response prediction. The model integrates 3D PET and 3D CT scans processed through separate 3D convolutional networks, along with laboratory biomarkers, via a concatenation-based fusion layer. The fused features are passed through fully connected layers to generate the prediction output. Interpretability is provided through backward analyses, including gradient maps, feature importance, and UMAP-based feature space analysis. Solid arrows represent forward pass data flow, dashed arrows backward pass dervied post hoc outputs. (Bottom) Overview of input modality combinations evaluated in our experiments.}
    \label{fig:model_and_table}
\end{figure}

We evaluated seven predictive models for PFS classification, exploring the predictive value of laboratory parameters, imaging, and multimodal data. First we trained a Random Forest classifier using only laboratory biomarkers. The laboratory biomarkers included are AST, ALT, CgA, and GGT. Subsequently, we developed two separate 3D convolutional neural networks (3D-CNNs) to predict PFS based on PET and CT imaging data, respectively. To investigate the benefit of multimodal integration, we constructed fusion models that combined (1) PET imaging with laboratory biomarkers, (2) CT imaging with laboratory biomarkers, and (3) PET and CT imaging with laboratory biomarkers. Finally, we assessed an advanced fusion model that integrates PET, CT, and laboratory  data, where the CT branch was initialized with a pretrained MONAI 3D CT segmentation model \cite{wasserthal2023totalsegmentator}, that we finetuned rather than being trained from scratch. Laboratory biomarkers were fused via concatenation to a flattened vector of image embeddings, and passed to 3 layers of MLPs. An overview of all our evaluated models is given in Figure \ref{fig:model_and_table}.
All models were trained using a learning rate of 0.01. We applied a dropout rate  of 0.1. Changing dropout to higher rates did not significantly change results. To prevent overfitting and employed the ADAM optimizer with a weight decay of 0.2. Again, different weight decay did not significantly influence results. We used Binary Cross Entropy Loss for all models. All imaging data underwent preprocessing prior to model training and statistical analysis. Visual artifacts were removed, voxel slopes were harmonized using dicom metadata, and finally the 3D volumes of all scans were normalized across training data. Each scan was subsequently resized to $75 \times 50 \times 50$ voxels. As our architectural design is fairly simple, our models are easily reproducable, given access to the data. 

\subsection{Statistical Analysis}
\label{sec:statistical_analysis}
To make sure our evaluation was reliable and not dependent on one specific data split, we used a repeated cross-validation approach. Therefore, we divided the dataset into three parts of equal size (3-fold cross-validation). In each round, two parts were used to train the model and the remaining part was used to test it. This rotation continued until each part had served once as the test set. We then repeated this entire 3-fold process five times, each time reshuffling the data. This repetition reduced the chance that the results were influenced by a particular way of splitting the data, giving us a more robust estimate of performance. Finally, we reported the average performance across all 5 repetitions, along with the standard error to indicate how much the results varied. Performance was measured using two widely applied metrics: the area under the receiver operating characteristic curve (AUROC), and the area under the precision–recall curve (AUPRC). To assess whether the observed performance differences between model families were statistically significant, we conducted nonparametric significance testing. Each model was assigned to one of three predefined groups based on its input modality: unimodal (PET-only, CT-only, or Random Forest on tabular data), one-image fusion (PET or CT combined with clinical variables), and dual fusion (both PET and CT combined, with or without pretraining). We first compared one-image fusion models against unimodal models, and then dual-fusion models against one-image fusion models. For each comparison, we used the Mann–Whitney U test (two-sided) to evaluate whether one group achieved higher performance than the other. To control for multiple testing, p-values were adjusted using the Bonferroni correction. In addition, we quantified the effect size with Cliff’s delta, which indicates how strongly one group tends to outperform the other. According to conventional interpretation, values of $|\delta| < 0.147$, $< 0.33$, $< 0.474$, and $\ge 0.474$ correspond to negligible, small, medium, and large effects, respectively.

\subsection{Model Analysis and Explainability}
To elucidate the decision-making process of our deep learning framework, we performed both \textit{representation space analysis} and \textit{biomarker relevance assessment} as well as qualitative explainability through \textit{saliency maps}. The aim was to provide transparency on how the model integrates multimodal information---namely, PET imaging and laboratory biomarkers---to arrive at its predictions.

\subsubsection{Umap Analysis of Feature Embeddings}
Uniform Manifold Approximation and Projection (UMAP)\cite{mcinnes2018umap} was employed to visualize the high-dimensional feature embeddings learned by the network. UMAP is a non-linear dimensionality reduction technique that projects complex feature spaces into two dimensions while preserving both local and global structural relationships. This makes it particularly suitable for identifying separable patient subgroups in latent space. 

\subsubsection{Feature Importance Analysis of Laboratory Biomarkers}
Feature importance analysis was performed to quantify the relative contribution of each laboratory biomarker to the predictive performance of different fusion strategies. For this, we evaluated the influence of gradients of laboratory values in the fusion layer on the model predictions.

\subsubsection{Qualitative Explainability}
Gradient-based saliency maps were computed to localize regions within the PET scans that most strongly influenced the model’s classification decisions. For each patient, voxel-wise gradients from the PET Fusion and PET Only models were backpropagated and mapped to the input PET volume to generate saliency heatmaps. These were overlaid on the original scans in three anatomical planes (axial, coronal, and sagittal) to visually highlight spatial patterns of model attention. 
\section{Results}
\subsection{Clinical Characteristics}
A total of 116 patients were included in the final study cohort with a median age of 66 years (range: 36--87). 41\% of patients were female. The most common primary sites were the small intestine (42\%) and pancreas (29\%), with 17\% of patients having a cancer of unknown primary (CUP). Most patients presented with hepatic metastases (73\%), frequently 
accompanied by lymphonodal or osseous spread. The majority of tumors were G2 (71\%) with a median Ki-67 index of 5\% (range: 1--40). Patients received a median of 3 PRRT cycles (range: 1--7). Key baseline laboratory values, including chromogranin~A (CgA) and $\gamma$-GT, are summarized in Table~\ref{tab:patient_characteristics}.

\begin{table}[]
\centering
\small
\begin{tabularx}{\textwidth}{
  >{\raggedright\arraybackslash}X
  >{\centering\arraybackslash}X
  >{\centering\arraybackslash}X
  >{\centering\arraybackslash}X
  >{\centering\arraybackslash}X}
\toprule
\textbf{Metric} & \textbf{Total} & \textbf{PFS $\leq$ 1 year} & \textbf{PFS $>1$ year} & \textbf{p-value} \\
\midrule
\multicolumn{5}{l}{\textbf{Patient Statistics}} \\
\midrule
Patient Count & 116 (100\%) & 42 (36\%) & 74 (64\%) &  \\
Age in years & 66 (36--87) & 66 (36--87) & 66 (36--80) & 0.945 \\
Male & 68 (59\%) & 23 (55\%) & 45 (61\%) & 0.560 \\
Female & 48 (41\%) & 19 (45\%) & 29 (39\%) & 0.560 \\
\midrule
\multicolumn{5}{l}{\textbf{Primary Location}} \\
\midrule
Small intestine & 49 (42\%) & 19 (45\%) & 30 (41\%) & 0.697 \\
Pancreas & 34 (29\%) & 8 (19\%) & 26 (35\%) & 0.090 \\
Colon/Rectum & 12 (10\%) & 4 (10\%) & 8 (11\%) & 1.000 \\
Stomach & 1 (1\%) & 0 (0\%) & 1 (1\%) & 1.000 \\
CUP & 20 (17\%) & 11 (26\%) & 9 (12\%) & 0.073 \\
\midrule
\multicolumn{5}{l}{\textbf{Metastatic Spread}} \\
\midrule
Hepatic & 85 (73\%) & 28 (67\%) & 57 (77\%) & 0.276 \\
Lymphonodal & 75 (65\%) & 26 (62\%) & 49 (66\%) & 0.689 \\
Osseous & 35 (30\%) & 12 (29\%) & 23 (31\%) & 0.836 \\
Peritoneal & 19 (16\%) & 6 (14\%) & 13 (18\%) & 0.796 \\
Pulmonal & 5 (4\%) & 1 (2\%) & 4 (5\%) & 0.652 \\
\midrule
\multicolumn{5}{l}{\textbf{Functionality}} \\
\midrule
Yes & 40 (34\%) & 19 (45\%) & 21 (28\%) & 0.072 \\
No & 75 (65\%) & 22 (52\%) & 53 (72\%) & 0.045 \\
Unknown & 1 (1\%) & 1 (2\%) & 0 (0\%) & 0.362 \\
\midrule
\multicolumn{5}{l}{\textbf{Grading}} \\
\midrule
G1 & 23 (20\%) & 8 (19\%) & 15 (20\%) & 1.000 \\
G2 & 82 (71\%) & 29 (69\%) & 53 (72\%) & 0.833 \\
G3 & 6 (5\%) & 2 (5\%) & 4 (5\%) & 1.000 \\
Unknown & 5 (4\%) & 3 (7\%) & 2 (3\%) & 0.351 \\
Ki67-Index \% & 5 (1--40) & 5 (1--25) & 5 (1--40) & 0.501 \\
\midrule
\multicolumn{5}{l}{\textbf{Laboratory Parameters}} \\
\midrule
\mbox{CgA~in~$\mu$g\slash l} & \mbox{419 (24--99590)} & \mbox{821 (25--99590)} & \mbox{262 (24--15100)} & 0.001 \\
AST in U\slash l & 28 (14--139) & 32 (14--123) & 28 (14--139) & 0.123 \\
ALT in U\slash l & 28 (7--132) & 27 (7--96) & 28 (10--132) & 0.774 \\
$\gamma$-GT in U\slash l & 61 (9--691) & 95 (21--688) & 50 (9--691) & 0.014 \\
De Ritis ratio & 1.12 (0.46--3.43) & 1.16 (0.46--3.43) & 1.07 (0.53--2.87) & 0.223 \\
PRRT Cycles & 4 (1--7) & 2 (1--4) & 4 (1--7) & $<0.001$ \\
\bottomrule
\end{tabularx}
\caption{Summary of patient characteristics stratified by PFS. Categorical variables were compared with Fisher's exact test; continuous variables with the Wilcoxon rank-sum (Mann--Whitney) test. Values are counts with percentages, or median (min--max). “Functionality” refers to the presence of clinically relevant hormone secretion by NETs. Parts of the presented cohort overlap with previously published studies \cite{10.3390/cancers13040635,wetz2023plasma}. See the \textit{Methods} section for further details.
}
\label{tab:patient_characteristics}
\end{table}

\subsection{Progression Free-Survival}
\begin{figure}[!h]
    \centering
    \includegraphics[width=0.7\textwidth]{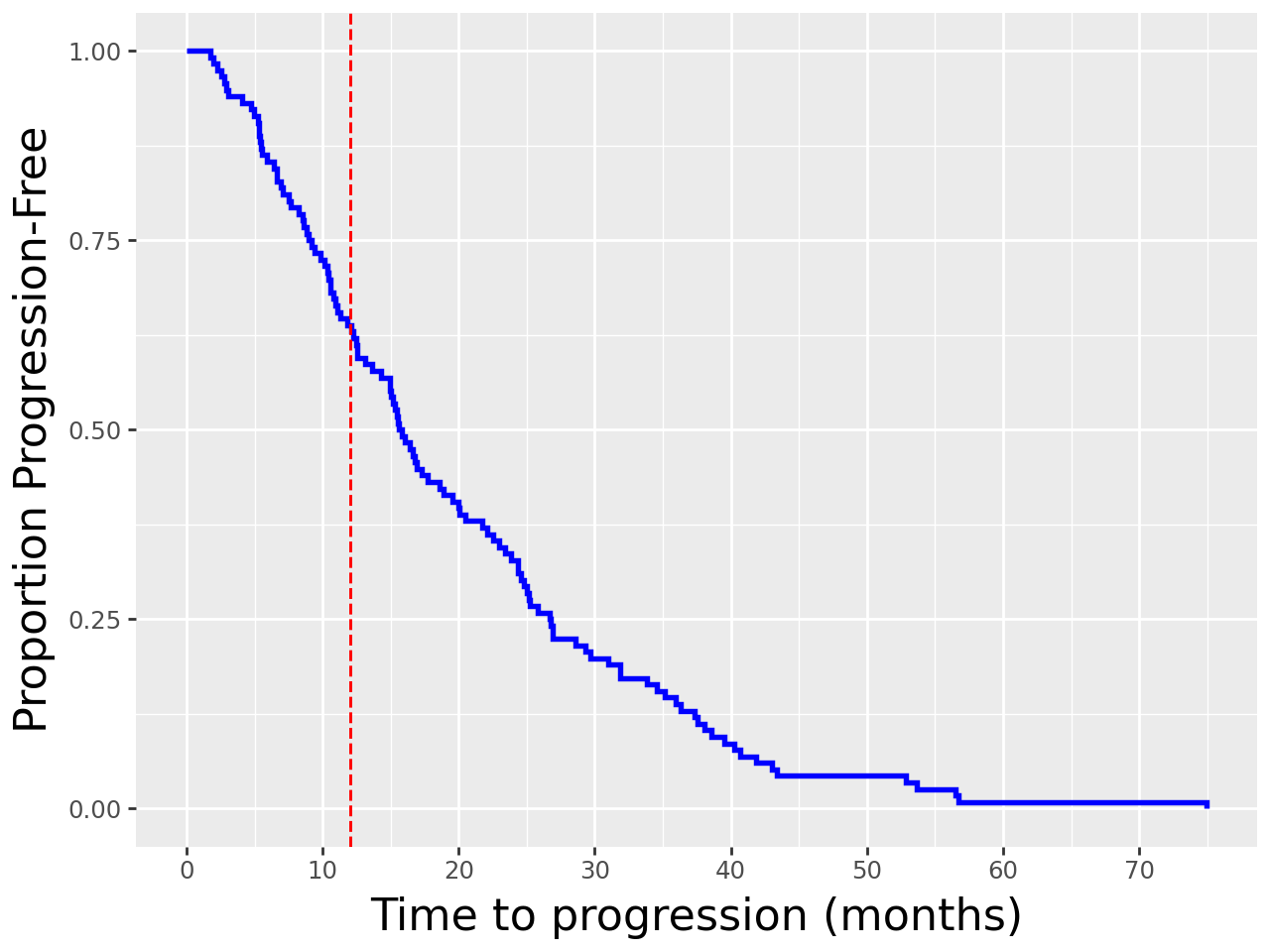}
    \caption{Kaplan-Meier curve for progression-free survival (PFS) in the total study cohort (n = 116) of patients with neuroendocrine tumors treated with [\textsuperscript{177}Lu]Lu-DOTATOC PRRT. Vertical dashed red line indicates our split into high and low therapy response. No censored patients are included, and all patients eventually had progress.}
    \label{fig:kaplan_meier_pfs}
\end{figure}

To evaluate treatment outcomes in the study cohort, we first analyzed the PFS. Figure~\ref{fig:kaplan_meier_pfs} illustrates the PFS distribution of the patient cohort. The median PFS for the total cohort was 15.7 months (interquartile range [IQR]: 9.1--26.7 months), indicating that half of the patients remained progression-free for at least this duration. No censored patients were included, as target values are necessary for a sample to be used in our deep learning setup. Our cohort includes only patients that eventually had a progress. Patients were stratified into short-PFS ($\leq$ 1 year) and long-PFS ($>$ 1 year) group.  When comparing these two subgroups most clinical characteristics did not differ significantly between the groups, including age, sex distribution, primary tumor location, metastatic pattern, tumor functionality, and histological grading (all $p > 0.05$). Notably, baseline chromogranin~A (CgA) level was significantly higher in patients with shorter PFS ($p = 0.003$), and elevated $\gamma$-GT levels ($p = 0.02$). In addition, patients with early progression had received fewer PRRT cycles ($p < 0.01$), which is expected, as treatment is typically discontinued in the event of disease progression prior to completion of the planned cycles. 

\subsection{Deep Learning Predictive Model for Progression-Free Survival}
\begin{figure}[!h]
    \centering
    \includegraphics[width=0.9\textwidth]{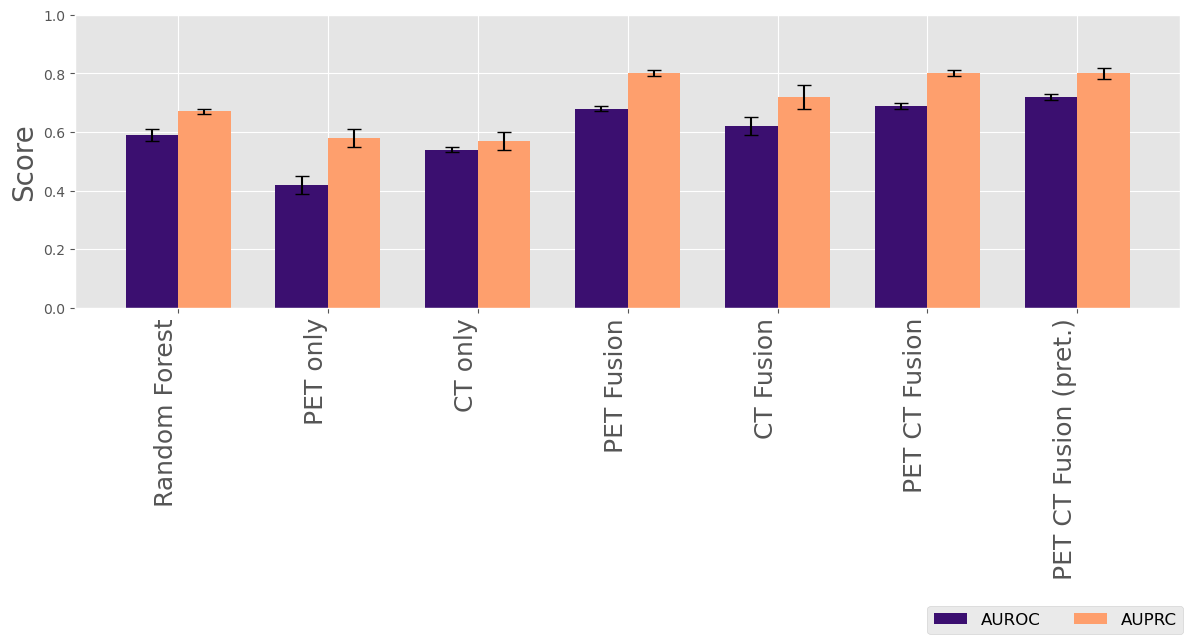}
    \caption{Barplot comparison of AUROC and AUPRC across models.}
    \label{fig:model_performance_barplot}
\end{figure}

\begin{figure}[!h]
    \centering
    \includegraphics[width=1\textwidth]{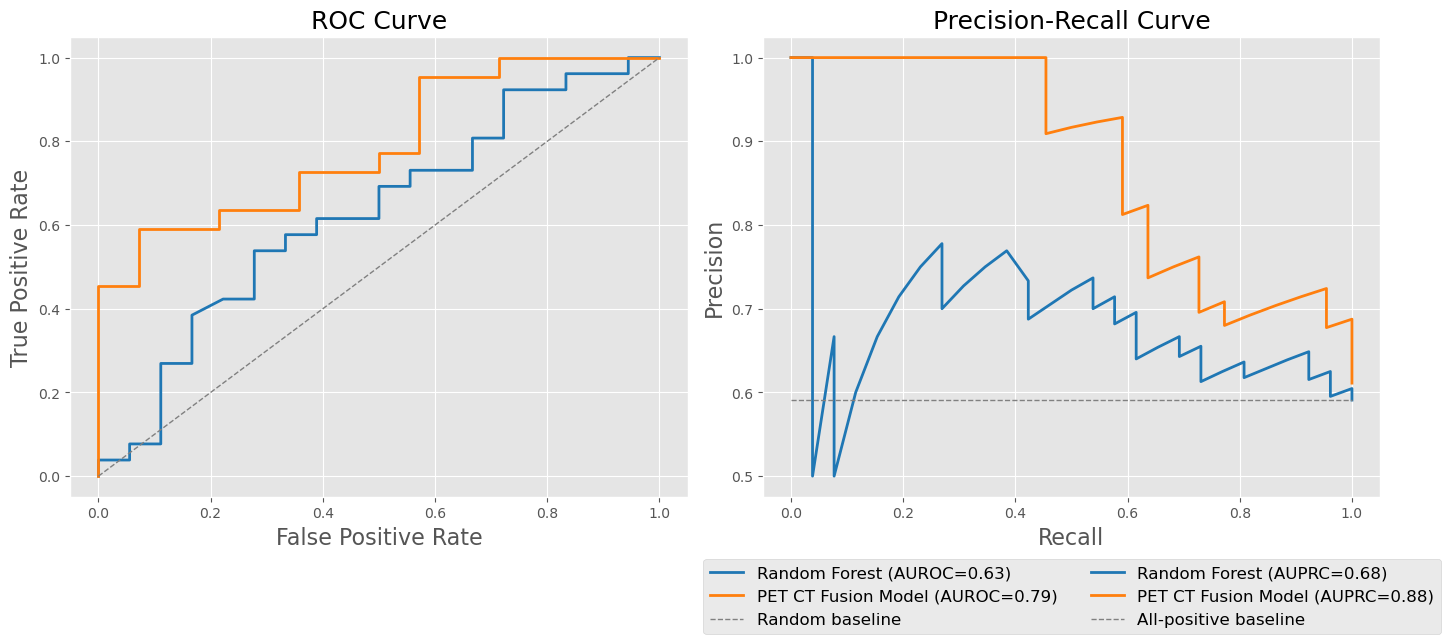}
    \caption{Comparison of predictive performance between the Random Forest baseline (laboratory values only) model and the PET CT Fusion model. The plot is showing the example of a single representative cv fold. Left: ROC curves showing True Positive Rate versus False Positive Rate; the dashed gray line represents a random baseline. Right: Precision-Recall curves illustrating the trade-off between precision and recall; the dashed gray line indicates the all-positive baseline. The PET CT Fusion model consistently outperforms the Random Forest baseline, as reflected in higher AUROC and AUPRC values. For cross-validation metrics, refer to Table \ref{tab:model_performance}.}
    \label{fig:auroc_auprc}
\end{figure}

\begin{figure}[!h]
    \centering
    \includegraphics[width=0.7\textwidth]{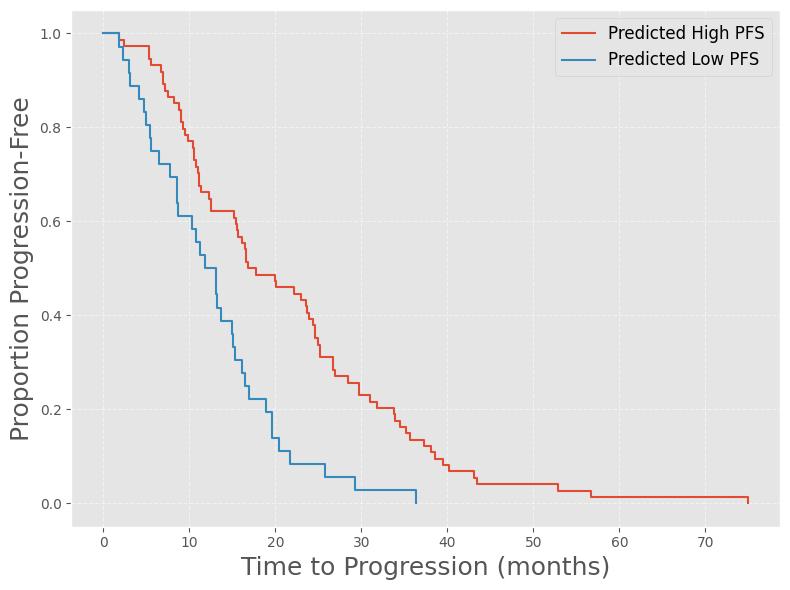}
    \caption{Kaplan-Meier Curve of study cohort, stratified by our model prediction output probabilities $\hat{y}$ (low PFS: $\hat{y} < 0.5$,  high PFS: $\hat{y} >= 0.5$). Note that as we did not include censored patients, and all patients in our cohort eventually had progression, the y axis represents the proportion of progression patients at a given time. Log-Rank Test: $p = 0.0001$.}
    \label{fig:kaplan_meier_stratified}
\end{figure}

We applied a series of unimodal and multimodal deep learning architectures to assess the added value of integrating imaging and laboratory data for predicting PFS. 
Our results demonstrate that integrating multiple data modalities consistently improves model performance in progression-free survival (PFS) classification (Table \ref{tab:model_performance}). The baseline Random Forest model, trained solely on laboratory biomarkers, achieved moderate performance (AUROC: $0.59 \pm 0.02$, AUPRC: $0.67 \pm 0.01$, Accuracy: $0.61 \pm 0.02$). In contrast, unimodal 3D convolutional neural networks trained on PET or CT data alone yielded lower discriminative performance, particularly for the PET-only model (AUROC: $0.42 \pm 0.03$). The CT-only model performed slightly better (AUROC: $0.54 \pm 0.01$), though both remained inferior to the model using only laboratory parameters. Introducing laboratory features into imaging-based models led to improvements: the PET-laboratory fusion model achieved an AUROC of $0.68 \pm 0.01$ and the highest accuracy overall ($0.65 \pm 0.01$), suggesting strong complementarity between PET imaging and laboratory data. Similarly, the CT-laboratory fusion model improved over the CT-only model across all metrics, reaching an AUROC of $0.62 \pm 0.03$. Further combining all three modalities—PET, CT, and laboratory data—resulted in additional performance gains. The PET-CT-laboratory fusion model achieved an AUROC of $0.69 \pm 0.01$ and matched the highest AUPRC score ($0.80 \pm 0.01$), reinforcing the value of multimodal integration. Finally, initializing the CT branch with a pretrained model further boosted the AUROC to $0.72 \pm 0.01$, indicating that leveraging pretrained representations can enhance predictive performance. Figure \ref{fig:model_performance_barplot} gives a visual overview of predictive performance of all models. For additional insights into predictive performance, Figure \ref{fig:auroc_auprc} displays ROC and Precision-Recall curves of a single exemplary cross validation fold for our PET CT Fusion model. We can clearly see the improved predictive performance compared to the Random Forest model and a random baseline. Statistical testing applied as described in \ref{sec:statistical_analysis} suggests that adding iteratively more modalities improves predictive model performance significantly ($p < 0.01$).
\newline
\newline
To further explore the clinical relevance of our model predictions, we performed a Kaplan–Meier analysis stratified by the model-derived probability of long progression-free survival (PFS) (see Figure \ref{fig:kaplan_meier_stratified}. The plot shows a representative single cross validation of the PET Fusion model. Patients predicted by our model to have a high therapy response (probability $\geq$ 0.5) demonstrated a markedly prolonged PFS compared with those in the low-response group (probability < 0.5).The median PFS was 17.25 months in the high-response group versus 12.43 months in the low-response group, corresponding to a statistically significant difference (log-rank p = 0.0001; test statistic 15.18). As all patients in our cohort eventually experienced progression, no censoring occurred, and the survival curves therefore display the proportion of patients who had progressed at a given time point. These findings indicate that our multimodal model is capable of clinically meaningful risk stratification, separating patients into distinct prognostic groups based solely on baseline data prior to therapy initiation.

\begin{table}[!h]
    \centering
    \begin{tabular}{l@{\hskip 1cm}c@{\hskip 3cm}c}
        \hline
        Model & AUROC & AUPRC \\
        \hline
        RF (Laboratory values only)  & $0.59 \pm 0.02$ & $0.67 \pm 0.01$ \\
        PET only & $0.42 \pm 0.03$ & $0.58 \pm 0.03$ \\
        CT only & $0.54 \pm 0.01$ & $0.57 \pm 0.03$ \\

        && \\
        
        PET Fusion & $0.68 \pm 0.01^{*}$ & $\mathbf{0.80 \pm 0.01^{*}}$ \\
        CT Fusion & $0.62 \pm 0.03^{*}$ & $0.72 \pm 0.04^{*}$ \\

        && \\
        
        PET CT Fusion & $0.69 \pm 0.01^{\dagger}$ & $\mathbf{0.80 \pm 0.01}$ \\
        PET CT Fusion (pretrained CT) & $\mathbf{0.72 \pm 0.01^{\dagger}}$ & $\mathbf{0.80 \pm 0.02}$ \\
        \hline
    \end{tabular}
    \caption{Performance metrics (AUROC and AUPRC) for all models. 
    Values are reported as mean~$\pm$~standard error. 
    Bolded values indicate the best performance in each metric. 
    Significance markers denote statistical improvement over the next lower model family: 
    $^{*}$~$p < 0.01$~(vs.~unimodal); 
    $^{\dagger}$~$p < 0.01$~(vs.~one-image fusion). 
    All significant differences correspond to large effect sizes (Cliff’s~$\delta > 0.8$).}
    \label{tab:model_performance}
\end{table}

\subsection{Results Model Analysis and Explainability}
\subsubsection{UMAP Analysis}
\begin{figure}[!h] % Requires \usepackage{float} in preamble
    \centering
    \includegraphics[width=\textwidth]{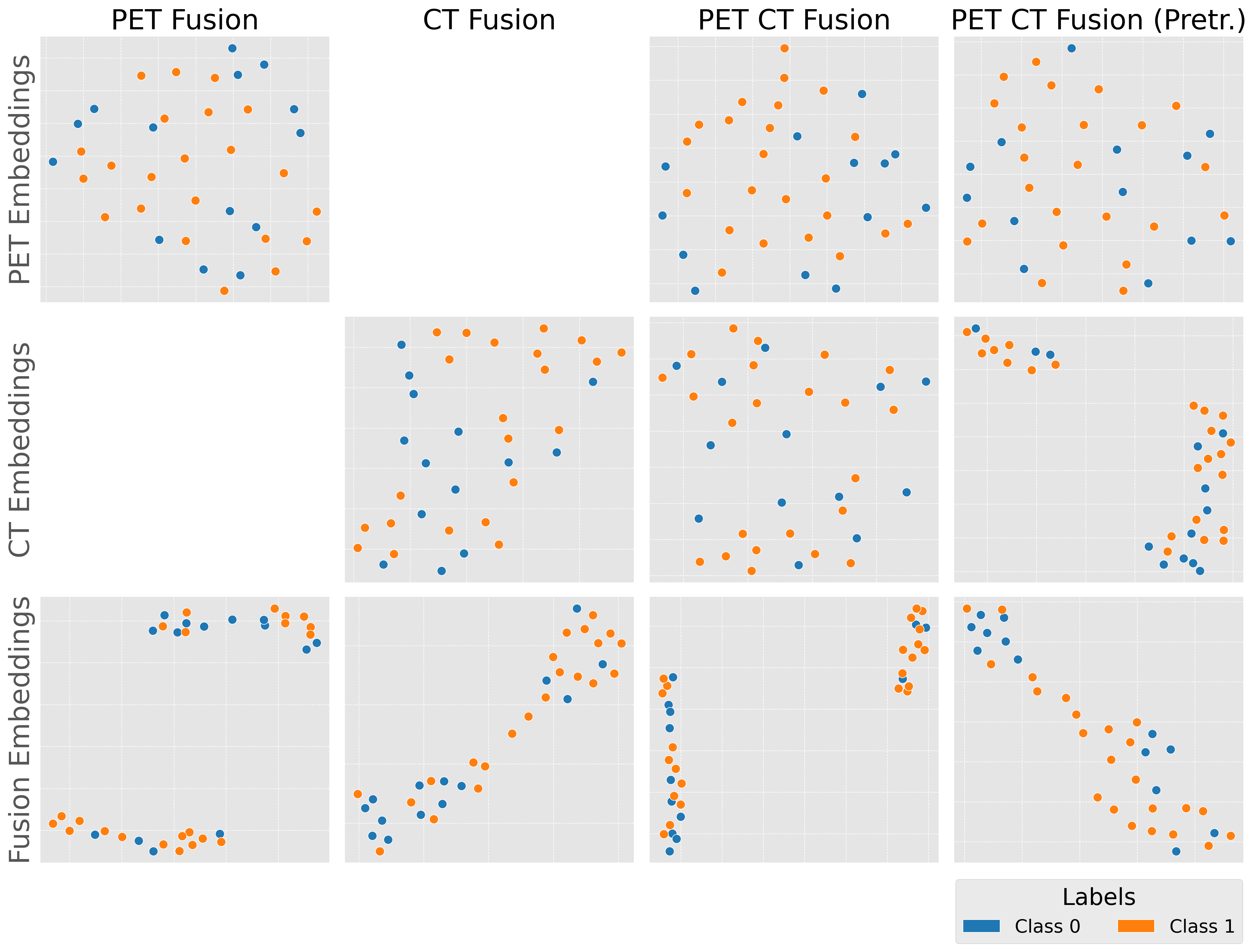}
    \caption{
    U-MAP projection of learned feature embeddings from different fusion strategies. 
    Row 1: Embeddings derived from PET imaging alone. Embeddings are mixed up between classes, therefore PET imaging alone is not capable of good discrimination.
    Row 2: Embeddings derived from CT imaging alone. Similar to PET imaging only, embeddings are scattered with no clear distinction. Notably, the pretrained CT model displays better clustered embeddings, due to prior exposure to CT imaging. 
    Row 3: Fusion embeddings combining PET, CT, and laboratory biomarkers reveal markedly improved class separation, with tighter intra-class clustering and clearer inter-class boundaries. This illustrates the synergistic effect of multimodal integration in capturing disease-related variation that is not apparent in single-modality embeddings.
    }
    \label{fig:umap}
\end{figure}

Figure~\ref{fig:umap} displays the UMAP projections of embeddings derived from different fusion strategies, PET imaging combined with laboratory biomarkers (PET Fusion), CT imaging combined with laboratory biomarkers (CT Fusion),  joint PET and CT imaging fused with laboratory biomarkers (PET-CT Fusion), and the same PET-CT fusion model with the CT branch initialized from a pretrained network (PET-CT Fusion, Pretrained). PET-derived embeddings alone show weak class separation. In contrast, the fused PET-CT embeddings---particularly when incorporating pretrained CT features---exhibit markedly improved clustering, with tighter intra-class grouping and greater inter-class separation. This suggests multimodal fusion enriches the feature space, enabling the model to capture subtle features associated with PRRT response. 

\begin{figure}[!h]
    \centering
    \includegraphics[width=\textwidth]{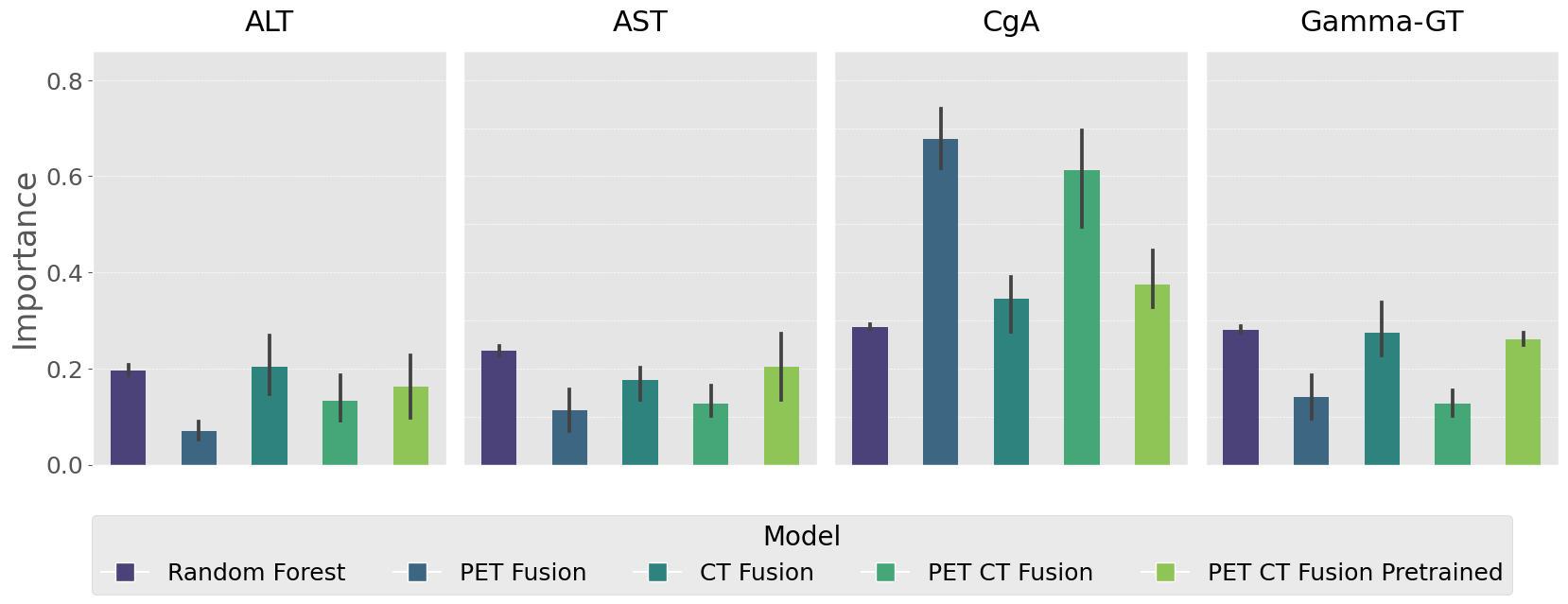} % Adjust width as needed
           \caption{
           Feature importance of selected laboratory biomarkers across different fusion strategies and a Random Forest baseline. Importance values were computed using a permutation-based approach, with higher values indicating stronger contribution to model predictions. Four biomarkers—ALT, AST, CgA, and Gamma-GT—consistently ranked among the most relevant features across all model configurations. CgA exhibited the highest importance in both PET-CT fusion models, suggesting a strong association with the target outcome when combined with imaging-derived features. 
        }

    \label{fig:feature} % Optional label for referencing
\end{figure}

\subsubsection{Feature Importance}
Our evaluation of feature importance of laboratory values is displayed in Figure \ref{fig:feature}. Across all model configurations, ALT, AST, CgA, and Gamma-GT consistently emerged as key discriminative variables. Notably, CgA exhibited the highest importance in both PET-CT fusion approaches, highlighting its strong association with the target outcome when combined with imaging- related features. The pretrained PET-CT fusion model generally preserved or enhanced biomarker relevance compared to the non-pretrained variant, suggesting that the integration of well-learned CT representations can strengthen the interpretive value of specific laboratory measures. The Random Forest baseline, while ranking the same biomarkers highly, demonstrated lower absolute importance values, underscoring the advantage of deep multimodal learning in capturing non-linear relationships between biochemical and imaging features.

\subsubsection{Qualitative Explainability}
\begin{figure*}[!h]
    \centering
    % --- Top subfigure: Placeholder image ---
    \begin{subfigure}[b]{\textwidth}
        \centering
        \includegraphics[width=\textwidth]{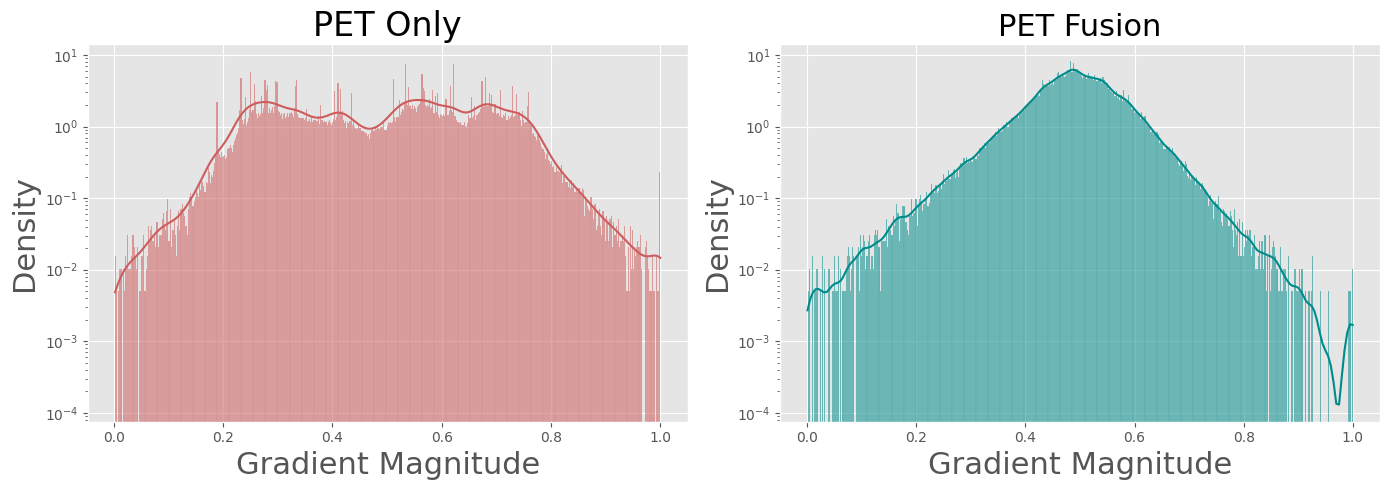}
    \caption{Comparison of gradient magnitude distributions between PET Only and PET Fusion models.}

        \label{fig:gradmaps_placeholder}
    \end{subfigure}
    
    \vspace{1.5em} % space between subfigures
    % --- Bottom subfigure: Table ---
    \begin{subfigure}[b]{1\textwidth}
        \centering
        \label{fig:gradmaps_table}
        \vspace{0.5em}
        \begin{tabular*}{\textwidth}{@{\extracolsep{\fill}} l c l}
            \hline
            \textbf{Metric} & \textbf{Value} & \textbf{Range \& Interpretation} \\
            \hline
            Wasserstein Distance & 0.090 & $[0, \infty)$ \hspace{0.5em} $\uparrow$ (9\% of range) \\
            KS Statistic (p $<$ 0.001) & 0.241 & $[0, 1]$ \hspace{0.5em} $\uparrow$ (moderate) \\
            Jensen--Shannon Divergence & 0.418 & $[0, 1]$ \hspace{0.5em} $\uparrow$ (moderate) \\
            Energy Distance & 0.183 & $[0, \infty)$ \hspace{0.5em} $\uparrow$ (18\% of range) \\
            Bhattacharyya Distance & 0.207 & $[0, \infty)$ \hspace{0.5em} $\uparrow$ (moderate) \\
            Histogram Overlap & 0.518 & $[0, 1]$ \hspace{0.5em} $\downarrow$ (52\% overlap) \\
            \hline
        \end{tabular*}
        \caption{Quantitative comparison of gradient distributions between PET      Only and PET Fusion models across multiple distance metrics.}
    \end{subfigure}
    
    \caption{Visual and quantitative comparison of global gradient distributions for PET only and PET Fusion models.
    (a) Histograms of global gradient distributions for both models. 
    (b) Quantitative comparison of global gradient distributions for both models.}
    \label{fig:gradients_global}
\end{figure*}

\begin{figure}[h!]
    \centering
    % --- New top full-width plot ---
    \begin{subfigure}[b]{\textwidth}
        \centering
        \includegraphics[width=\textwidth]{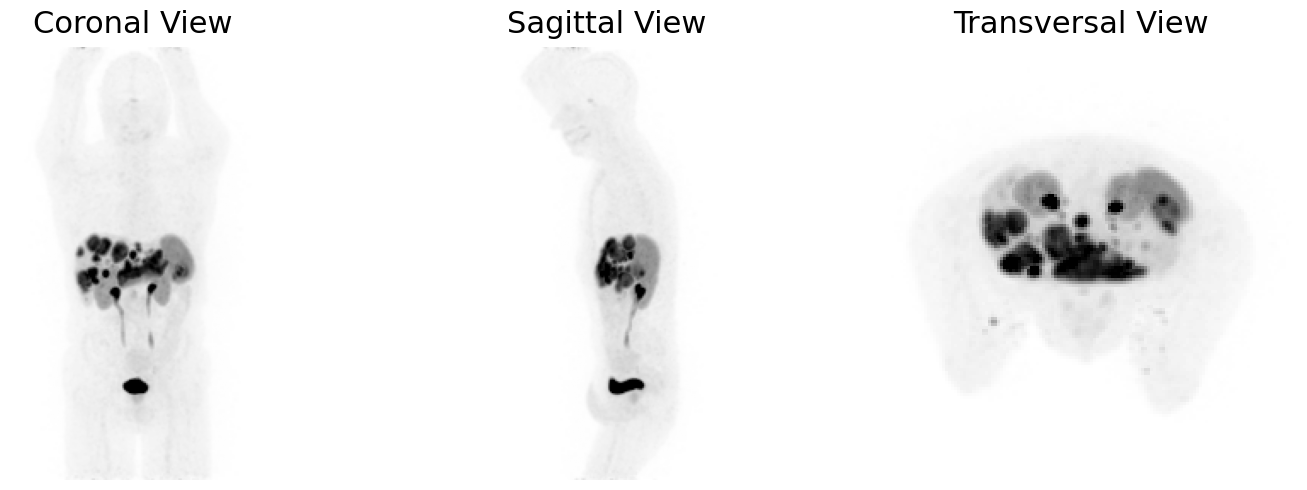}
        \caption{
        Example of a three-dimensional \({}^{68}\)Ga-DOTA-PET scan. 
        }
        \label{fig:raw_pet_example}
    \end{subfigure}

    \vspace{0.8em} % small spacing

    %--- First row: Gradient heatmaps ---
    \begin{subfigure}[b]{0.43\textwidth}
        \centering
        \includegraphics[width=\textwidth]{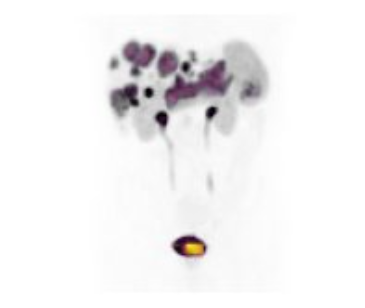}
        \caption{Gradient heatmap PET Only.}
        \label{fig:grad_pet_only}
    \end{subfigure}
    \hfill
    \begin{subfigure}[b]{0.43\textwidth}
        \centering
        \includegraphics[width=\textwidth]{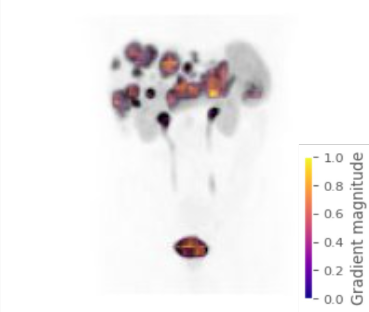}
        \caption{Gradient heatmap PET Fusion.}
        \label{fig:grad_pet_fusion}
    \end{subfigure}

    %--- Minimal vertical spacing ---
    \vspace{0.0em}

    %--- Second row: Gradient histograms ---
    \begin{subfigure}[b]{0.48\textwidth}
        \centering
        \includegraphics[width=\textwidth]{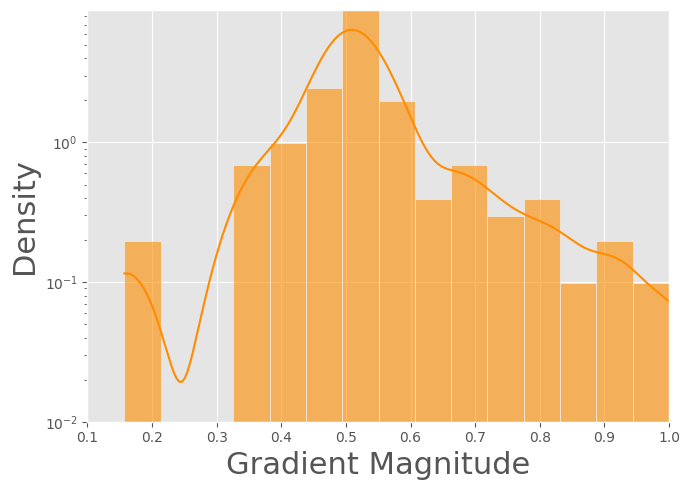}
        \caption{Gradient distribution for PET Only.}
        \label{fig:hist_pet_only}
    \end{subfigure}
    \hfill
    \begin{subfigure}[b]{0.48\textwidth}
        \centering
        \includegraphics[width=\textwidth]{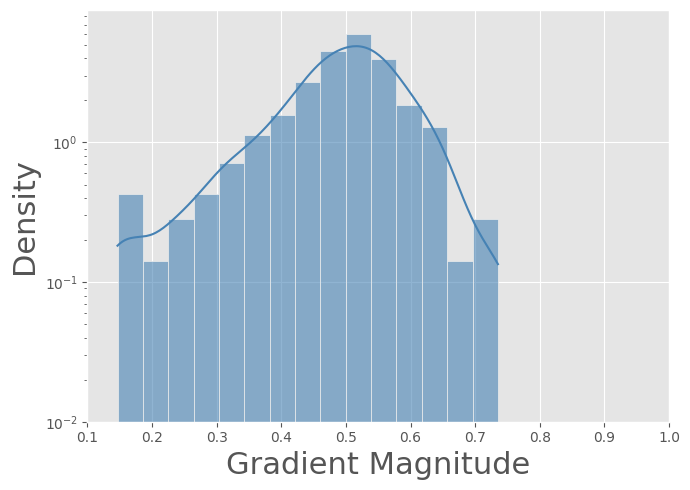}
        \caption{Gradient distribution for PET Fusion.}
        \label{fig:hist_pet_fusion}
    \end{subfigure}

    %--- Main caption ---
    \caption{
        Example PET scan (a) and comparison of gradient maps for a single sample between PET Only and PET Fusion models. 
        Panels (b)--(c) show the gradient heatmap overlays, while (d)--(e) show the corresponding gradient distributions. 
        Gradient maps were filtered with $v_\mathrm{min}=0.3$, omitting smaller gradients.
    }
    \label{fig:pet_fusion_vs_pet_only_sample}
\end{figure}

Figures \ref{fig:gradients_global} and \ref{fig:pet_fusion_vs_pet_only_sample} show the results of our qualitative explainability analysis. Firstly, we analyzed the global distributions of gradient magnitudes across all test sets of an entire cross validation (Figure~\ref{fig:gradients_global} (a)) for PET Only and PET Fusion models. The PET Only model exhibits an irregular, noisy distribution with substantial density fluctuations across the gradient range. In contrast, PET Fusion displays a smooth, unimodal distribution centered around moderate gradient magnitudes. Notably, PET Only shows a pronounced shift toward higher gradient values (spanning the full [0, 1] range with significant density beyond 0.6), which—coupled with the model’s poor generalization performance—is indicative of exploding gradients, a well-known phenomenon in deep learning that can hinder stable learning and meaningful representation formation~\cite{glorot2010understanding,bengio1994learning,hochreiter1991untersuchungen,ceni2025random}. Quantitative comparisons using multiple statistical distance metrics confirmed substantial divergence between the distributions of the two models: Wasserstein Distance (0.090), Kolmogorov–Smirnov Statistic (0.241, $p < 0.001$), Jensen–Shannon Divergence (0.418), Energy Distance (0.183), Bhattacharyya Distance (0.207), and Histogram Overlap (0.518) (Figure~\ref{fig:gradients_global}(b)). We further illustrate these differences at the individual-case level. An example of a raw PET scan is shown in Figure~\ref{fig:pet_fusion_vs_pet_only_sample} (a). Figure~\ref{fig:pet_fusion_vs_pet_only_sample} (b) and (c) compare the corresponding saliency maps of the PET Only and PET Fusion models for the same patient. Brighter colors indicate voxels with stronger contributions to the model’s prediction. The PET Fusion model predominantly focuses on relevant tumorous regions, while the PET Only model assigns high importance to the bladder. In addition to visual inspection, we compared the gradient magnitude distributions of both models (Figure~\ref{fig:pet_fusion_vs_pet_only_sample} (d) and (e)). The PET Only model exhibits numerous large gradients and an irregular, fragmented distribution, whereas the PET Fusion model produces a smoother, more coherent gradient distribution with fewer extreme values. In total, our gradient analyses support our earlier findings from model performance (Table~\ref{tab:model_performance}) and internal feature representations (Figure~\ref{fig:umap}): the PET Fusion model learns more physiologically meaningful signal patterns associated with PRRT effectiveness, while the PET Only model fails to capture relevant information. The contrasting distributional shapes underscore fundamental differences in training stability: PET Fusion’s concentrated, bell-shaped profile reflects well-regulated gradient flow, while PET Only’s diffuse, erratic pattern signals optimization instability.

\section{Discussion}

% Main findings/summary
In this study, we developed and evaluated a multimodal deep learning model integrating somatostatin receptor PET, CT imaging, and laboratory biomarkers to predict progression-free survival in patients undergoing [\textsuperscript{177}Lu]Lu-DOTATOC PRRT. Our results show that unimodal imaging models alone — whether based on SR-PET or CT — were insufficient to provide clinically meaningful predictive performance and, in fact, performed worse than the model using only laboratory data. In contrast, combining complementary imaging modalities with laboratory biomarkers in a fusion architecture substantially enhanced predictive accuracy and robustness. Importantly, we incorporated explainability into the model by leveraging three-dimensional gradient maps and biomarker relevance analyses, enabling  interpretation of decision-driving features.
\newline
\newline
% Comparison with existing literature
As reported previously, analyses of the cohort characteristics showed that baseline levels of CgA and gamma-GT were higher in patients with shorter PFS \cite{10.3390/cancers13040635}. Consistent with this finding, earlier studies have demonstrated an inverse association between baseline CgA and clinical outcome in NET patients undergoing PRRT \cite{10.2967/jnumed.118.224386}. A similar pattern was observed for gamma-GT, which has been linked to hepatic tumor burden and poorer prognosis in NET \cite{schmidt2024does}, and which in our analysis was consistently higher in patients with early progression. Interestingly, in the explainability analysis of our multimodal model, CgA also emerged as a parameter with notable contribution to prediction (see Figure 4). This suggests that despite its limited value as a stand-alone biomarker — being strongly influenced by non-tumor–related factors such as proton-pump inhibitor therapy, renal dysfunction, or other comorbidities — CgA can still provide complementary prognostic information when integrated with imaging and other laboratory features.
Although previous authors have applied machine learning approaches to predict the prognosis of patients with NETs \cite{jiang2023predicting,gao2025machine}, these models have typically relied on single data modalities, predominantly clinical and laboratory data. For example, Jiang et al.\ used deep learning on population-based data from the SEER registry—containing only demographic, clinical, and pathological variables—to predict survival in pancreatic NETs \cite{jiang2023predicting}. Likewise, Gao et al.\ developed a machine learning model for prognosis estimation in gastroenteropancreatic NET patients with liver metastases using solely clinical parameters \cite{gao2025machine}.
\newline
In addition to clinical and laboratory parameters, imaging information has increasingly been explored as a means of predicting outcomes after PRRT in NET patients. SR-PET/CT provide essential information on tumor burden and receptor expression, and several groups have investigated whether quantitative or radiomic features could be used for prognostic modeling. For example, in a recent study, Opali\'nska et al. found that a significant decrease in liver-normalized SUV\textsubscript{max} in NET lesions on [\textsuperscript{68}Ga]Ga-DOTA-TATE PET/CT following PRRT was associated with a lower risk of disease progression over a 20-month follow-up \cite{opalinska2022potential}. This suggests that PET/CT-derived SUV\textsubscript{lmax} in NET lesions may serve as an additional and independent predictor of treatment outcome.
Further, Laudicella et al. reported that the [\textsuperscript{68}Ga]Ga-DOTA-TATE PET/CT radiomic features HISTO\_Skewness and HISTO\_Kurtosis predicted PRRT response for individual lesions of both primary and metastatic GEP-NETs, regardless of tumor origin, with AUCs of 0.745 and 0.722, respectively \cite{laudicella202268ga}. Importantly, in the CLARINET trial, Pavel et al.\ reported that deep learning models based on CT imaging alone failed to outperform conventional laboratory markers such as chromogranin A and specific growth rate (SLDr) \cite{pavel2023use}. Similarly, in our study, the model based solely on CT scans or SR-PET showed no meaningful prognostic value and performed worse than a baseline model using laboratory biomarkers. Only when laboratory and imaging data were combined in a multimodal fusion model did we observe a relevant increase in predictive performance. These results further underscore the complementary nature of PET and CT imaging in capturing distinct yet clinically relevant aspects of disease biology. While SR-PET emphasizes functional and metabolic activity, CT provides higher resolution anatomical detail. 
From a clinical perspective, this implies that radiomic signatures from combined PET and CT imaging---augmented by biochemical markers---may reflect pathological differences more accurately than any modality alone. 
When integrated within a shared feature space alongside laboratory biomarkers, the combined modality seems to offer a richer and more complete representation of patient status. 
This multimodal synergy enables the network to detect patterns that may be too subtle to discern in either modality alone, thereby improving the robustness and generalizability of the learned representations. 
\newline
\newline
% Strengths and limitations
We recognize several limitations of our study. First, the sample size was relatively small, which raises concerns about the robustness and generalizability of the findings. Training deep learning models on limited data can lead to overfitting; although we employed cross-validation and regularization techniques, a larger dataset would be needed to ensure the model’s performance is consistent and not an artifact of our particular cohort. Second, our analysis was retrospective. This inherently carries risks of selection bias (e.g. only patients who completed PRRT were included) and confounding factors that prospective studies could better control.
A key limitation of our study is the absence of an external validation cohort, which restricts the generalizability of our findings. Nonetheless, the PET/CT images were acquired using different scanner systems, potentially introducing variability due to differences in reconstruction algorithms. Given that all required inputs—laboratory parameters as well as SR-PET and CT imaging—are routinely obtained as part of the standard diagnostic work-up in patients scheduled for PRRT, prospective validation in larger multicenter cohorts appears feasible.
Another methodological limitation concerns the heterogeneity of CT acquisition protocols. Among the 116 patients included, 42 underwent non-contrast CT, whereas the remaining patients received contrast-enhanced whole-body scans acquired during the venous contrast phase. This ensured complete anatomical coverage and optimal alignment between PET and CT images. However, arterial phase imaging could have improved the visualization of certain lesions, especially hepatic metastases, and might have enhanced the accuracy of image-based analyses. A further limitation of our study is that we considered only a relatively small fraction of the potentially available clinical information. Additional data such as genetic profiles, advanced laboratory parameters, histological images, or multiplex staining might have provided further predictive value. At the same time, novel biomarkers such as the NETest are gaining increasing attention. Future models that integrate such high-specificity biomarkers with deep learning predictions could further enhance the accuracy and clinical utility of prognostic tools in NET patients undergoing PRRT.
\newline
\newline
In contrast to many previous studies in this field, our work makes a contribution with respect to explainability, moving beyond the paradigm of “black-box” deep learning models. While gradient-based visualization did highlight tumor regions, as expected, it also consistently emphasized areas such as kidneys, spleen, and urinary bladder. In line with the observed AUROC of 0.42 for the SR-PET–only model, these findings indicate that SR-PET data alone did not provide predictive value for PFS. Consistent with the inferior predictive performance, PET-only models produced noisier and less structured gradient maps, with strong activations concentrated in medically irrelevant regions (Figure \ref{fig:pet_fusion_vs_pet_only_sample}). In contrast, the PET Fusion model—though not entirely free of spurious correlations, which are expected to some extent in any explainability method—yielded clearer, more coherent gradient patterns that tend to focus more on clinically relevant tumorous regions. In general, saliency in non-tumor regions likely reflects a mix of relevant and spurious correlations inherent to the imaging data. Importantly, these correlations are not necessarily harmful in our setting: predictive performance emerges only after fusion with laboratory features, as supported by our UMAP analysis of embedding space, suggesting that the model leverages clinically meaningful interactions rather than relying solely on non-medically relevant image cues.
\newline
\newline
% Clinical implications and future directions
From a clinical perspective, the ability to stratify patients by likely PFS has significant implications. PRRT is an expensive and resource-intensive therapy, and not without toxicity, therefore, optimizing patient selection is critical. If a model identifies a patient as high risk for early progression, clinicians might consider adapting the treatment strategy. Such patients could benefit from closer monitoring during therapy and earlier response evaluation. The multimodal deep learning framework presented in this study builds on routinely available laboratory and imaging data, which may facilitate integration into interdisciplinary tumor board discussions and clinical workflows. Moreover, the architecture is designed to flexibly incorporate additional data sources in the future, such as genetic profiling, thereby further enhancing its predictive potential.

\bibliographystyle{unsrt}
\newpage
\bibliography{references}
\section*{Statements and Declarations}

\subsection*{Funding}
This work was supported by the Senate of Berlin and the
European Commision’s Digital Europe Programme (DIGITAL) as grant TEF-Health (101100700). Johannes Eschrich is a participant in the BIH Charité Junior Digital Clinician Scientist Program funded by the Charité – Universitätsmedizin Berlin, and the Berlin Institute of Health at Charité. 

\subsection*{Competing Interests}
The authors have no relevant financial or non-financial interests to disclose.

\subsection*{Author Contributions}

All authors contributed to the conception and design of the study.  
Data collection was performed by Tristan Ruhwedel, Zuzanna Kobus, Gergana Lishkova and Johannes Eschrich.  
Imaging evaluation was supervised by Julian M.\,M. Rogasch, Christoph Wetz, and Holger Amthauer.  
Model development and computational analysis were conducted by Simon Baur, Ekin Böke, Jackie Ma, and Wojciech Samek.  
Manuscript review and editing were performed by Christoph Roderburg, Frank Tacke, Holger Amthauer, Christoph Wetz, Henning Jann, Julian M.\,M. Rogasch, Jackie Ma, and Wojciech Samek.  
Study conception, clinical oversight, and supervision were provided by Johannes Eschrich.  
The first draft of the manuscript was written by Johannes Eschrich and Simon Baur.  
All authors read and approved the final version of the manuscript.

\subsection*{Data Availability}
The datasets generated and analysed during the current study are available from the corresponding author on reasonable request. Due to institutional and ethical restrictions, data are not publicly available.

\subsection*{Ethics Approval}
This study was performed in accordance with the ethical standards of the institutional research committee and with the 1964 Helsinki Declaration and its later amendments. Approval was granted by the Ethics Committee of Charité – Universitätsmedizin Berlin (Approval No.: EA1/016/23; Date: 24 February 2023).

\subsection*{Consent to Participate}
Informed consent was obtained from all individual participants included in the study.

\subsection*{Consent to Publish}
Not applicable. This manuscript does not contain any individual person’s data in any form (including individual details, images, or videos); therefore, consent for publication was not required.

\end{document}